# Computational optimization of convolutional neural networks using separated filters architecture

Elena Limonova[1], Alexander Sheshkus[2] and Dmitry Nikolaev[3]

[1]Moscow Institute of Physics and Technology, Institutskiy per. 9, Dolgoprudny, 141700, Russia
elena.e.limonova@gmail.com
[2]Smart Engines Ltd., Prospekt 60-letiya Octyabrya 9, Moscow, 117312, Russia
astdcall@gmail.com
[3]Institute For Information Transmission Problems (Kharkevich Institute), Bolshoy Karetny per. 19-1, Moscow, 127051, Russia
dimonstr@iitp.ru

## Abstract

This paper considers a convolutional neural network transformation that reduces computation complexity and thus speedups neural network processing. Usage of convolutional neural networks (CNN) is the standard approach to image recognition despite the fact they can be too computationally demanding, for example for recognition on mobile platforms or in embedded systems. In this paper we propose CNN structure transformation which expresses 2D convolution filters as a linear combination of separable filters. It allows to obtain separated convolutional filters by standard training algorithms. We study the computation efficiency of this structure transformation and suggest fast implementation easily handled by CPU or GPU. We demonstrate that CNNs designed for letter and digit recognition of proposed structure show 15% speedup without accuracy loss in industrial image recognition system. In conclusion, we discuss the question of possible accuracy decrease and the application of proposed transformation to different recognition problems. convolutional neural networks, computational optimization, separable filters, complexity reduction

## 1 Introduction

Neural networks are powerful and flexible tool of modern cognitive technologies, which can be applied to computer vision, classification or data processing problems. In a number of these problems convolutional neural networks (CNNs) are the most appropriate choice. Although CNNs provide high recognition accuracy, they can sometimes be very computationally demanding and it may be challenging to solve real-time classification problems on mobile platforms (e.g. credit card expiration date recognition [1]) or use them in real-time object detection systems [2, 3], where number of classifier executions per frame are enormous. It is worth noting that industrial recognition systems often have limited performance, strong memory and power restrictions and in some cases do not include GPUs.

All these circumstances lead to the necessity of high computational efficiency of neural networks. One of the most time consuming parts of CNN processing is convolution computation. One way to increase its performance is to use linear combination of separable filters in convolutions. Application of separable filters requires less calculations than application of standard 2D filters. Nevertheless, precise obtaining convolution as a linear combination of separable filters is impossible for already trained CNN with non-separable 2D filter. There are special methods to get the approximate separable representation, however they require additional computational effort and may cause accuracy loss [4, 5].

We propose the CNN structure transformation that represents 2D convolution as a linear combination of separable filters directly which means it can be trained on the same data with the same neural network training algorithm as the original CNN. During training the best solution of this kind can be found. It can be better than the result obtained by approximation of already trained 2D filter because this filter may be non-separable although the problem has another solution with separable filter.

The experiments showed that proposed structure transformation can speedup recognition in industrial image recognition system without accuracy loss. We conducted experiments on ARM processor Samsung Exynos 5422 with CNNs designed for letter and digit recognition.

The paper is organized as follows. In the Section 2 we describe convolutional layers organization and give computational complexity for standard and separable layer models. The 3rd Section contains information about the proposed CNN structure transformation. In the Section 4 we discuss efficient implementation of our CNN that can be used both for CPU or GPU, while in the 5th Section we demonstrate experimental results. Finally, the 6th Section gives conclusion and discussion.

**Related work.** The problem of improving performance of neural network processing is considered in a relatively small number of works. Some of currently considered methods are the usage of fixed point arithmetic [6, 7], approximation of 2D convolution filter by a set of separable filters [4, 5]. The fixed point arithmetic can speedup image recognition by 40% for image recognition with 16-bit quantization [6], and by the factor of 3 for speech recognition with 8-bit quantization at no cost in accuracy [7]. In [4] the authors use low rank approximation and clustering of filters. It resulted into 1.6 speedup of convolutional layers with an 11% increase in error rate. Rigamonti et al. [5] demonstrate how convolutional filters can be approximated by a set of separable 1D filters, getting processing performance increase without accuracy loss.

There are also hardware-specific methods of speedup: CPU or GPU oriented high-performance implementations. For example, the authors of [8] propose fast GPU implementation of separable filters. In [9, 10] effective neural network implementations suitable for mobile devices and embedded systems are described.

We suggest CNN structure transformation that uses separability and has effective implementation on CPU and GPU. Moreover, the performance can be additionally improved by any other method, for example, fixed point arithmetic.

## 2 Classical convolutional layer

This Section contains information about basic operations of convolutional layers of neural networks. Convolutional layers operate on relatively small areas of the input image and are able to extract local features of this area. The input of the layer is a multichannel image. The application of one convolutional filter can be described by the following expression:

$$O(x,y) = \sum_{\Delta x} \sum_{\Delta y} \sum_{c} I(x+\Delta x, y+\Delta y, c) w(\Delta x, \Delta y, c),$$

where $(x,y)$ is the point of the output, $c$ is the channel number, $O$ is output of convolution, $I$ is input image, $w$ is filter's matrix.

Each filter $w$ is multichannel, because it contains individual coefficients for each channel.

Since convolutional layer usually contains several filters to extract different features, the output is also multichannel image. After the convolution the bias is added and the non-linear activation function is applied. For example, such activation function can be rectifier or hyperbolic tangent. Let $K \times K$ be the size of each filter, $N \times M$ – the size of the input image, $L$ – the number of filters, $C$ – the number of input channels. Then the computation complexity will be $O(K^2 NMCL)$.

Now we move to the filter group concept which is supported for example by CUDA ConvNet [11]. The main idea is to divide input channels in groups and apply individual set of filters to each group with fewer number of channels. It allows to process different extracted features separately. The example is block sparse convolutional layer from CUDA ConvNet in Fig. 1.

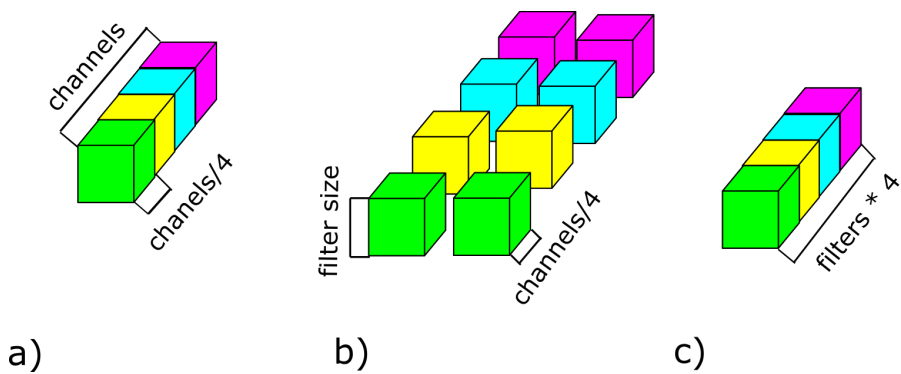

a)           b)           c)

Figure 1: Block sparse convolutional layer with 4 groups and 2 filters for each group [11].
    a) Input divided into 4 groups.
    b) 4 sets of filters each with four times less number of channels.
    c) Output produced by convolution of like-coloured filters and input channels.

## 3 CNN structure transformation

In this Section we describe CNN structure with separable filters. It is shown in Fig. 2. $L$ $C$-channel filters of standard CNN are replaced by the structure of 2 filters and fusing convolution $1 \times 1$. Let us look into it.

The steps are:

1. Apply $L$ $C$-channel filters of size $K \times 1$.
2. Divide channels into $L$ groups and apply $1$-channel filter $1 \times K$ to each group. Output of this step contains $L$ channels – one for each group.
3. Apply $L$ convolutions of size $1 \times 1$ to get linear combinations of step 2 outputs.
4. Apply non-linear activation function.

The complexity of such layer calculation is $O(NML(KC + K + L))$.

The number of weights is $KCL + KL + L$, which means that we have less weights than in standard convolutional layer.

With proposed convolution structure we can find linear combination of separable filters by standard training algorithm and it can be better than the one got by approximation methods. We approximate the space of effective filters for current problem by the space of separable filter combinations and search the solution in it. The separable approximation of any 2D filter lays also in this space and training algorithm can find it or better result if it exists. The particular common 2D filter obtained by training process may be non-separable even with modified training algorithm, therefore direct usage of the proposed CNN structure may give higher accuracy. Since a number of recognition problems allow separable approximation of filters almost without accuracy loss, it gives reasons to believe that our CNN structure does not cause accuracy decrease in wide range of problems. In Section 5 we demonstrate, that letter and digit recognition problems can be efficiently solved with the proposed CNN structure.

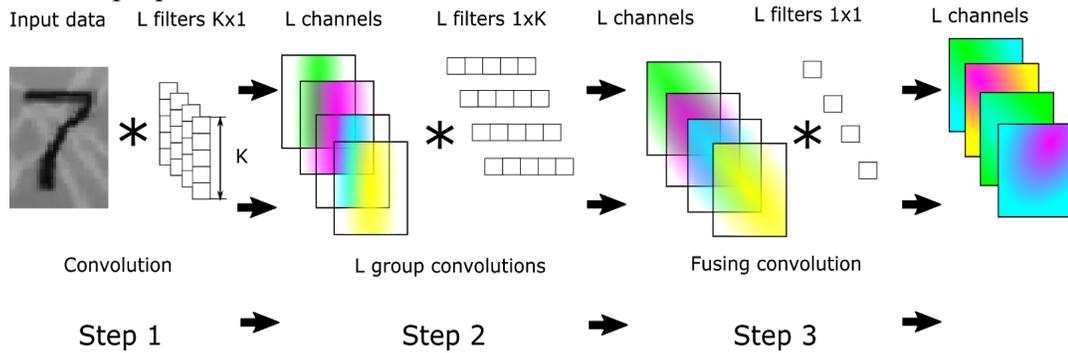

Figure 2: Introduced CNN structure.

## 4 Effective implementation

This Section is about effective implementation of convolution with groups on CPU. The standard convolution has fast implementation in matrix form [12] which can be performed with optimized BLAS packages. We can use this implementation on the step 1 of our convolution. However the step 2 can not be represented with such matrix operations directly. We suggest to describe this step with 3 dimensional structure (Fig. 3a). For the first matrix the dimensions are: the number of filters in the set, coefficients of group filters and groups. The second matrix is formed from input data and has the following dimensions: coefficients to convolve with filter, number of filter applications and groups. The result is also 3-dimensional: the number of filters in the set, the output for each group and the groups. The operation to get the result can be understood as matrix multiplication with vectors as elements. Vector length is equal to the number of groups and vector coefficients are the corresponding coefficients for different groups. Multiplication of these vectors is defined as coefficient-wise vector multiplication, while addition is conventional vector addition (Fig. 3b).

## 5 Experiments

The experiments were conducted on Samsung Exynos 5422 CPU. Matrix operations were implemented with the help of Eigen [13], which is fast and reliable C++ linear algebra library. It includes efficient CPU implementation of matrix operations for Intel x86 SSE, ARM NEON based on the compiler intrinsics. 16-bit fixed point arithmetic was used for additional performance gain. To train CNNs with separable filters we modified CUDA ConvNet because the basic version supports only square filters and even number of channels in a group.

We took 2 CNNs trained with Stochastic Gradient Descent method. They were designed to extract digits and letters from credit card images [1]. The input images had 1 channel and the size of 14x20 pixels. We used 90% of images for training set and 10% for test set to control recognition accuracy. The size of training set was more than million of samples. The comparison of the accuracy obtained by standard CNN structure and the proposed is represented in Table 1. The first column refers to the normal CNN with 8 $5\times 5$ filters, the second - to CNN with 8 $5\times 1$ and $1\times 5$ filters, the third — to CNN with 8 $5\times 1$ and $1\times 5$ plus 8 $1\times 1$ fusing convolutions. The experiments show that the proposed structure with fusing convolution gives practically the same recognition accuracy as the standard one. The results without fusing are much worse.

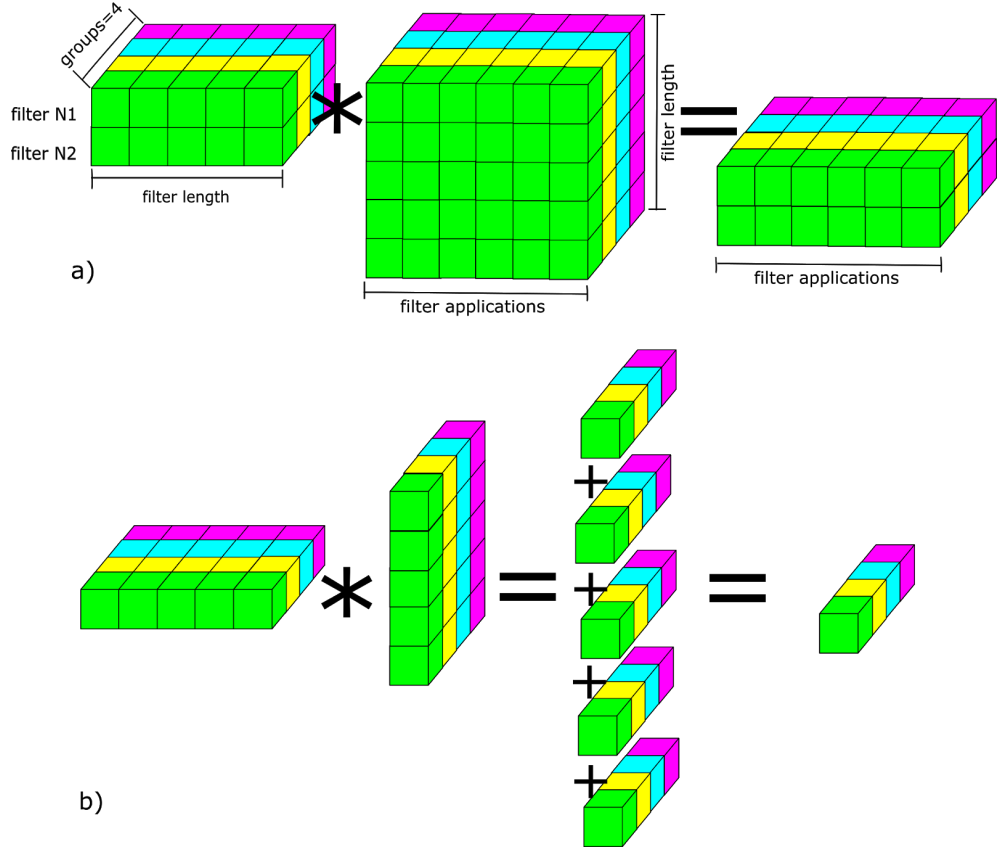

Figure 3: 3D representation of convolution with groups of filters. In this example we have 4 groups with 1-channel filters $1\times 5$, 6 filter applications and 2 filter sets.

a) Representation with the help of matrix multiplication, where elements of matrices are vectors. Vectors contain corresponding coefficients for different groups. Here vector length is 4.

b) Calculation of one output matrix element. Multiplication of matrix elements is defined as coefficient-wise vector multiplication, addition - as vector addition.

Table 1: Comparison of the error rate between classical and proposed CNN structures

| Alphabet | Standard CNN | Proposed CNN (without fusing) | Proposed CNN (with fusing) | Dataset size |
|---|---|---|---|---|
| Digits | 0.015% | 0.050% | 0.02% | $1.2\cdot 10^6$ |
| Roman letters | 0.065% | 0.19% | 0.065% | $1.8\cdot 10^6$ |

The usage of the proposed CNN structure transform reduces the number of multiplications in convolutional layer by the factor of about $1.39$. For CNN with one convolutional layer it resulted in 15% additional speedup of the processing after applied 16-bit quantization [6] which is a significant value for industrial recognition system with real time application. For CNNs with more filters or with filters of bigger sizes the result would be better.

# 6 Conclusion & Discussion

In this paper we introduced CNN structure transformation that makes CNN more computationally efficient due to the separability of convolution. We approximate the space of effective filters for recognition problems by the space of linear combinations of separable filters and search the solution by standard training software without algorithmic modifications. It means that convolutional layer is represented as a linear combination of separable filters with the help of filter groups concept, which allows to get separable convolution filters by standard methods. Moreover, there are reasons to think that proposed CNN structure can be used in a number of recognition problems without accuracy loss based on the existing research in this sphere. We demonstrate that image recognition accuracy does not decrease with digit and letter recognition as the example.

We suggested fast CPU implementation that can also be successfully used on GPU. It resulted in 15% speedup of CNN processing in comparison with well tuned BLAS implementation on CPU. However, convolution calculation was about $1.39$ times more efficient than the standard version. The speedup value is implementation dependent and can vary in different systems. Nevertheless, the complexity of computations of linear combination of separable filters is comparable with all the training or approximation methods. It means that the 15% speedup is specific to our system and CNN used, although separability of filters can give performance increase by the factor of 2.5 [14] or 30 [5] for other recognition problems.

This result can be important for deep neural networks processing that can be very time consuming or real-time recognition problems. It allows to improve performance of CNNs keeping recognition accuracy on the same level for a wide range of problems.

# 7 Acknowledgments

This work is supported by Russian Foundation for Basic Research (projects 15-29-06083 and 16-07-01167).